\documentclass{article} %
\usepackage{iclr2023_conference,times}
\pdfoutput=1

\usepackage{amsmath,amsfonts,bm}

\def\eqref#1{equation~\ref{#1}}

\def\1{\bm{1}}

\DeclareMathAlphabet{\mathsfit}{\encodingdefault}{\sfdefault}{m}{sl}
\SetMathAlphabet{\mathsfit}{bold}{\encodingdefault}{\sfdefault}{bx}{n}

\usepackage{hyperref}
\usepackage{url}

\usepackage[pdftex]{graphicx} 
\usepackage{booktabs}
\usepackage{xspace}
\usepackage{xcolor}
\usepackage{wrapfig}
\usepackage{xfrac}
\usepackage{tabularx}
\usepackage{tikz}
\usepackage{multirow}
\usepackage[mathscr]{euscript}
\usepackage{enumitem}
\definecolor{natural}{rgb}{0.7137,0.3333,0.3333}
\definecolor{specialized}{rgb}{0.4118,0.6431,0.4314}
\definecolor{structured}{rgb}{0.3254,0.4431,0.6666}
\definecolor{all}{rgb}{0.7529,0.4902,0.6471}

\definecolor{alexey}{rgb}{0.8, 0.0, 0.8}

\definecolor{matthias}{rgb}{0.0, 0.8, 0.8}

\definecolor{sylvain}{rgb}{0.8, 0.8, 0.0}

\newcommand{\oursattn}{SSA\xspace}
\newcommand{\oursabbrv}{Spikformer\xspace}
\newcommand{\oursfull}{Spiking Transformer\xspace}

\usepackage{amsfonts}       % blackboard math symbols
\usepackage{nicefrac}       % compact symbols for 1/2, etc.
\usepackage{microtype}      % microtypography
\usepackage{xcolor}         % colors
\usepackage{xspace}
\usepackage{multirow}
\usepackage{subfigure}
\usepackage{adjustbox}
\usepackage{pifont}
\usepackage{bbding}
\usepackage{fontawesome} 
\newcommand{\cmark}{\ding{51}}%
\newcommand{\xmark}{\ding{55}}%
\newcommand{\tabincell}[2]{\begin{tabular}{@{}#1@{}}#2\end{tabular}}
\usepackage{tikz}
\newcommand*{\circled}[1]{\lower.7ex\hbox{\tikz\draw (0pt, 0pt)%
    circle (.5em) node {\makebox[1em][c]{\small #1}};}}
\title{{Spikformer: When Spiking Neural Network Meets Transformer }}

\author{
{\normalsize
$^{1,2}$Zhaokun Zhou \ \ \ $^{1,2}$Yuesheng Zhu$^{*}$ \ \ \ $^{4}$Chao He \ \ \ $^{2}$Yaowei Wang \ \ \ $^3$Shuicheng Yan
\vspace{0.3mm}
} \\
{\normalsize
\ \textbf{$^{1,2}$Yonghong Tian \ \ \ $^{1,2}$Li Yuan}\thanks{Indicates the corresponding author.}\vspace{0.3mm}
}\\
\ $^1$Peking University \qquad $^2$Peng Cheng Laboratory \qquad $^3$Sea AI Lab \\ \ $^4$ Shenzhen EEGSmart Technology Co., Ltd.  \vspace{0.7mm} \\
{\ \ \texttt{\{yuanli-ece\}@pku.edu.cn}} \\
}

\iclrfinalcopy %
\begin{document}

\maketitle
\begin{abstract}
We consider two biologically plausible structures, the Spiking Neural Network (SNN) and the self-attention mechanism.
The former offers an energy-efficient and event-driven paradigm for deep learning,
%such as the currently mainstream ConvNet architecture, e.g., Spiking ResNet. The self-attention mechanism with 
while the latter has the ability to capture feature dependencies, enabling Transformer to achieve good performance.
%, which is also an important feature of the human biological systems. 
It is intuitively promising to explore the marriage between them.
%However, the direct use of vanilla self-attention does not make full use of the characteristics of SNNs. 
%quanhong: no proof/analysis support
In this paper, we consider leveraging both self-attention capability and biological properties of SNNs, and propose a novel Spiking Self Attention (\oursattn) as well as a powerful framework, named \oursfull (\oursabbrv). 
The \oursattn mechanism in \oursabbrv models the sparse visual feature by using spike-form Query, Key, and Value without softmax. Since its computation is sparse and avoids multiplication, \oursattn is efficient and has low computational energy consumption. It is shown that \oursabbrv with \oursattn can outperform the state-of-the-art SNNs-like frameworks in image classification on both neuromorphic and static datasets. \oursabbrv (66.3M parameters) with comparable size to SEW-ResNet-152 (60.2M, 69.26\%) can achieve $74.81\%$ top1 accuracy on ImageNet using 4 time steps, which is the state-of-the-art in directly trained SNNs models. Codes will be avaiable at \href{https://github.com/ZK-Zhou/spikformer}{\oursabbrv}.

% Due to simple addition operations and decomposability of SSA,
% \oursattn achieves low computational complexity and low computational energy consumption. 
% , whose computation consists of pure addition operations.
% which is the first above $7\%$ among all directly trained SNNs models.

% Extensive experiments show our \oursfull achieves competitive performance compared to the state-of-the-art SNNs-like frameworks. For example, 

\end{abstract}

\section{Introduction}
 As the third generation of neural network \citep{maass1997networks}, the Spiking Neural Network (SNN) is very promising for its low power consumption, event-driven characteristic, and biological plausibility \citep{roy2019towards}. 
With the development of artificial neural networks (ANNs), SNNs are able to lift performance by borrowing advanced architectures from ANNs, such as ResNet-like SNNs \citep{hu2018residual,fang2021deep, zheng2021going, hu2021advancing}, Spiking Recurrent Neural Networks \citep{lotfi2020long} and Spiking Graph Neural Networks \citep{zhu2022graph}.
Transformer, originally designed for natural language processing \citep{vaswani2017attention},
has flourished for various tasks in computer vision, including image classification \citep{dosovitskiy2020image,yuan2021tokens}, object detection \citep{carion2020end,zhu2020deformable,liu2021swin}, semantic segmentation \citep{wang2021pyramid,yuan2021volo} and low-level image processing \citep{chen2021pre}. Self-attention, the key part of Transformer, selectively focuses on information of interest, and is also an important feature of the human biological system \citep{whittington2022relating,caucheteux2022brains}. Intuitively, it is intriguing to explore applying self-attention in SNNs for more advanced deep learning, considering the biological properties of the two mechanisms.

\begin{figure}[t!]
\begin{center}
\includegraphics[width=\textwidth]{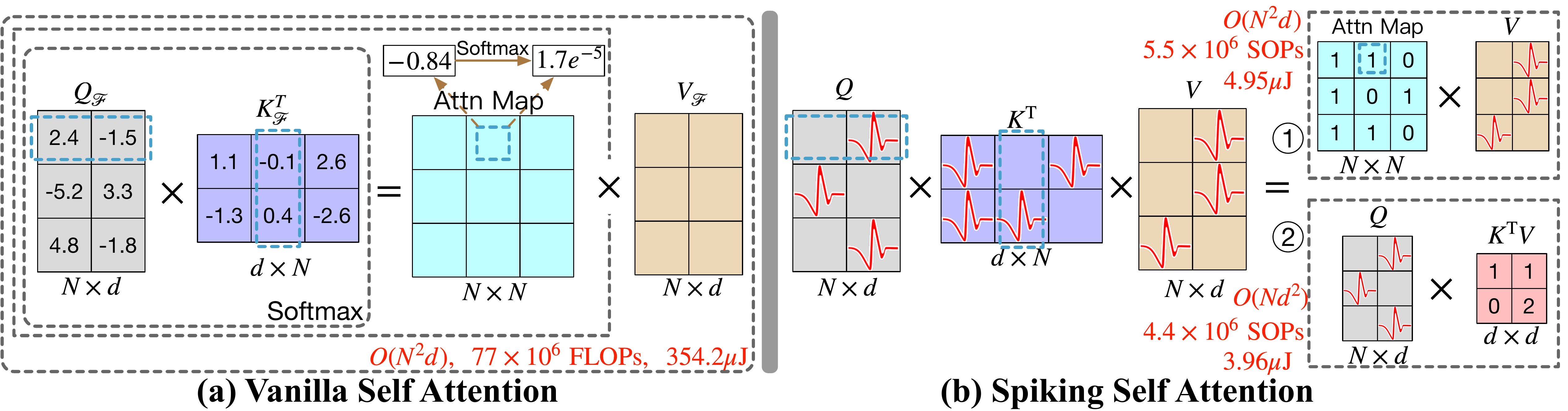}
\vspace{-14pt}
\end{center}
\caption{\small Illustration of vanilla self-attention (VSA) and our Spiking Self Attention (\oursattn). A red spike indicates a value of 1 at that location. The blue dashed boxes provide examples of matrix dot product operation. For convenience, we choose one of the heads of \oursattn, where $N$ is the number of input patches and $d$ is the feature dimension of one head. $\rm{FLOPs}$ is the floating point operations and $\rm{SOPs}$ is the theoretical synaptic operations. The theoretical energy consumption to perform one calculation between Query, Key and Value in one time step is obtained from $8$-encoder-blocks 512-embedding-dimension \oursabbrv on ImageNet test set according to \citep{kundu2021hire, hu2018residual}. More details about the calculation of theoretical SOP and energy consumption  are included in appendix. \ref{sec:energy}. (a) In VSA, $Q_{\mathcal{F}},K_{\mathcal{F}},V_{\mathcal{F}}$ are float-point forms. After the dot-product of $Q_{\mathcal{F}}$ and $K_{\mathcal{F}}$, the softmax function regularizes negative values in the attention map to positive values. (b) In \oursattn, all value in attention map is non-negative and the computation is sparse using spike-form $Q, K, V$ ($5.5\times 10^6$ VS. $77 \times 10^6$ in VSA). Therefore, the computation in \oursattn consumes less energy compared with VSA ({$354.2\mu \rm{J}$}). In addition, the \oursattn is decomposable (the calculation order of $Q,K$ and $V$ is changeable).
}
\vspace{-4mm}
\label{fig:ssa}
\end{figure}

It is however non-trivial to port the self-attention mechanism into SNNs. 
In vanilla self-attention (VSA) \citep{vaswani2017attention}, there are three components: Query, Key, and Value. As shown in Figure \ref{fig:ssa}(a), standard inference of VSA is firstly obtaining a matrix by computing the dot product of float-point-form Query and Key; then softmax, which contains exponential calculations and division operations, is adopted to normalize the matrix to give the attention map which will be used to weigh the Value.
The above steps in VSA do not conform to the calculation characteristics of SNNs, i.e., avoiding multiplication. 
Moreover, the heavy computational overhead of VSA almost prohibits applying it directly to SNNs. 
Therefore, in order to develop Transformer on SNNs, we need to design a new effective and computation-efficient self-attention variant that can avoid multiplications.

%To address the above limitations, 
We thus present Spiking Self Attention (\oursattn), as illustrated in Figure \ref{fig:ssa}(b). \oursattn introduces self-attention mechanism to SNNs for the first time, which models the interdependence using spike sequences. In \oursattn, the Query, Key, and Value are in spike form which only contains of $0$ and $1$. 
The obstacles to the application of self-attention in SNNs are mainly caused by softmax. 
1) As shown in Figure \ref{fig:ssa}, the attention map calculated from spike-form Query and Key has natural non-negativeness, which ignores irrelevant features. Thus, we do not need the softmax to keep the attention matrix non-negative, which is its most important role in VSA \citep{qin2022cosformer}. 
2) The input and the Value of the \oursattn are in the form of spikes, which only consist of 0 and 1 and contain less fine-grained feature compared to the float-point input and Value of the VSA in ANNs. So the float-point Query and Key and softmax function are redundant for modeling such spike sequences.
Tab. \ref{tab: ablation1} illustrates that our \oursattn is competitive with VSA in the effect of processing spike sequences. 
Based on the above insights, we discard softmax normalization for the attention map in \oursattn. Some previous Transformer variants also discard softmax or replace it with a linear function. For example, in Performer \citep{choromanski2020rethinking}, positive random feature is adopted to approximate softmax; CosFormer \citep{qin2022cosformer} replaces softmax with ReLU and cosine function.

With such designs of \oursattn, the calculation of spike-form Query, Key, and Value avoids multiplications and can be done by logical AND operation and addition. Also, its computation is very efficient. Due to sparse spike-form Query, Key and Value (shown in appendix \ref{sec:fr}) and simple computation, the number of operations in \oursattn is small, which makes the energy consumption of \oursattn very low. 
Moreover, our \oursattn is decomposable after deprecation of softmax, which further reduces its computational complexity when the sequence length is greater than the feature dimension of one head, as depicted in Figure \ref{fig:ssa}(b) \ding{172} \ding{173}.

Based on the proposed \oursattn, which well suits the calculation characteristics of SNNs, we develop the \oursfull (\oursabbrv).
%, which serves as a backbone in image classification. 
An overview of \oursabbrv is shown in Figure \ref{fig:model}. 
It boosts the performance trained on both static datasets and neuromorphic datasets. To the best of our knowledge, it is the first time to explore the self-attention mechanism and directly-trained Transformer in the SNNs. To sum up, there are three-fold contributions of our work:
% Compared with the original patch splitting using large kernel convolution layer or linear layer, we found that multi-layer spike-form convolution stem would bring much better performance.
\begin{itemize}[leftmargin=*]
	\item We design a novel spike-form self-attention named Spiking Self Attention (\oursattn) for the properties of SNNs. Using sparse spike-form Query, Key, and Value without softmax, the calculation of \oursattn avoids multiplications and is efficient. 
 \vspace{-2mm}
% 	It thus achieves low computational cost and low computational energy consumption.
	\item We develop the \oursfull(\oursabbrv) based on the proposed \oursattn. To the best of our knowledge, this is the first time to implement self-attention and Transformer in SNNs. 
	\item Extensive experiments show that the proposed architecture outperforms the state-of-the-art SNNs on both static and neuromorphic datasets. It is worth noting that we achieved more than $74\%$ accuracy on ImageNet with $4$ time steps using directly-trained SNN model for the first time.
\end{itemize}

\section{Related Work}
\textbf{Vision Transformers.}  For the image classification task, a standard vision transformer (ViT) includes a patch splitting module, the transformer encoder layer(s), and linear classification head. The Transformer encoder layer consists of a self-attention layer and a multi perception layer block.  Self-attention is the core component making ViT successful. By weighting the image-patches feature value through the dot-product of query and key and softmax function, self-attention can capture the global dependence and interest representation \citep{katharopoulos2020transformers,qin2022cosformer}. 
Some works have been carried out to improve the structures of ViTs. Using convolution layers for patch splitting has been proven to be able to accelerate convergence and alleviate the data-hungry problem of ViT \citep{xiao2021early,hassani2021escaping}. There are some methods aiming to reduce the computational complexity of self-attention or improve its ability of modeling visual dependencies \citep{song2021ufo,yang2021focal,rao2021dynamicvit,choromanski2020rethinking}. This paper focuses on exploring the effectiveness of self-attention in SNNs and developing a powerful spiking transformer model for image classification.

\textbf{Spiking Neural Networks.}  
Unlike traditional deep learning models that convey information using continuous decimal values, SNNs use discrete spike sequences to calculate and transmit information. Spiking neurons receive continuous values and convert them into spike sequences, including the Leaky Integrate-and-Fire (LIF) neuron \citep{wu2018spatio}, PLIF \citep{fang2021incorporating}, etc. There are two ways to get deep SNN models: ANN-to-SNN conversion and direct training.
{In ANN-to-SNN conversion \citep{cao2015spiking,hunsberger2015spiking,rueckauer2017conversion,bu2021optimal,meng2022training,wang2022signed}, the high-performance pre-trained ANN is converted to SNN by replacing the ReLU activation layers with spiking neurons.} The converted SNN requires large time steps to accurately approximate ReLU activation, which causes large latency \citep{han2020rmp}. In the area of direct training, 
SNNs are unfolded over the simulation time steps and trained in a way of backpropagation through time \citep{lee2016training,shrestha2018slayer}. Because the event-triggered mechanism in spiking neurons is non-differentiable, the surrogate gradient is used for backpropagation \citep{lee2020enabling,neftci2019surrogate}. {\cite{xiao2021training} adopts implicit differentiation on the equilibrium state to train SNN.} Various models from ANNs have been ported to SNNs. However, the study of self-attention on SNN is currently blank. \cite{yao2021temporal} proposed temporal attention to reduce the redundant time step. {\cite{zhang2022spiking,zhang2022spike} both use ANN-Transformer to process spike data, although they have 'Spiking Transformer' in the title. \cite{mueller2021spiking} provides a ANN-SNN conversion Transformer, but remains vanilla self-attention which does not conform the characteristic of SNN.} In this paper, we will explore the feasibility of implementing self-attention and Transformer in SNNs.

As the fundamental unit of SNNs, the spike neuron receives the resultant current and accumulates membrane potential which is used to compare with the threshold to determine whether to generate the spike.  We uniformly use LIF spike neurons in our work. The dynamic model of LIF is described as:
{\setlength{\abovedisplayskip}{3pt}
\setlength{\belowdisplayskip}{3pt}
\begin{align}
&H[t] = V[t-1] + \frac{1}{\tau}\left(X[t] - (V[t-1]-V_{reset})\right),\\
&S[t] = \Theta(H[t]-V_{th}),\\
&V[t]=H[t]~(1-S[t]) + V_{reset}S[t],
\end{align}}
where $\tau$ is the membrane time constant, and $X[t]$ is the input current at time step $t$. When the membrane potential $H[t]$ exceeds the firing threshold $V_{th}$, the spike neuron will trigger a spike $S[t]$. $\Theta(v)$ is the Heaviside step function which equals 1 for $v\geq 0$ and 0 otherwise. $V[t]$ represents the membrane potential after the trigger event which equals $H[t]$ if no spike is generated, and otherwise equals to the reset potential $V_{reset}$.
% Note that we have omitted the simulation time steps $T$ of spike neuron in the rest paper for brevity.

% , such as image classification \citep{touvron2021deit,yuan2021tokens,touvron2021going}, object detection \citep{carion2020end,zhu2020deformable}, semantic segmentation \citep{wang2021pyramid,liu2021swin}, image generation \citep{parmar2018image} and video task \citep{zhou2018end,zeng2020learning,arnab2021vivit}.
% \cite{hu2018residual,fang2021deep, zheng2021going} mainly study the residual structure on SNNs. There are some works on combination of recurrent neural networks (RNNs) and SNNs \citep{lotfi2020long,yin2020effective}. Spiking GNN is proposed for graph representation \citep{zhu2022graph}.

% Transformers are originally designed for natural language processing \citep{vaswani2017attention,brown2020language}. After DETR \citep{carion2020end} and ViT \citep{dosovitskiy2020image}, vision transformer models have become a new backbone in various visual tasks.
\newcommand{\op}[1]{\operatorname{#1}}
\newcommand{\mbf}[1]{\mathbf{#1}}
\vspace{-4mm}
\section{Method}
\vspace{-4mm}
{
\setlength{\belowcaptionskip}{3pt}
\begin{figure}[t]
\begin{center}
\begin{tabular}{c}
\includegraphics[width=.85\textwidth]{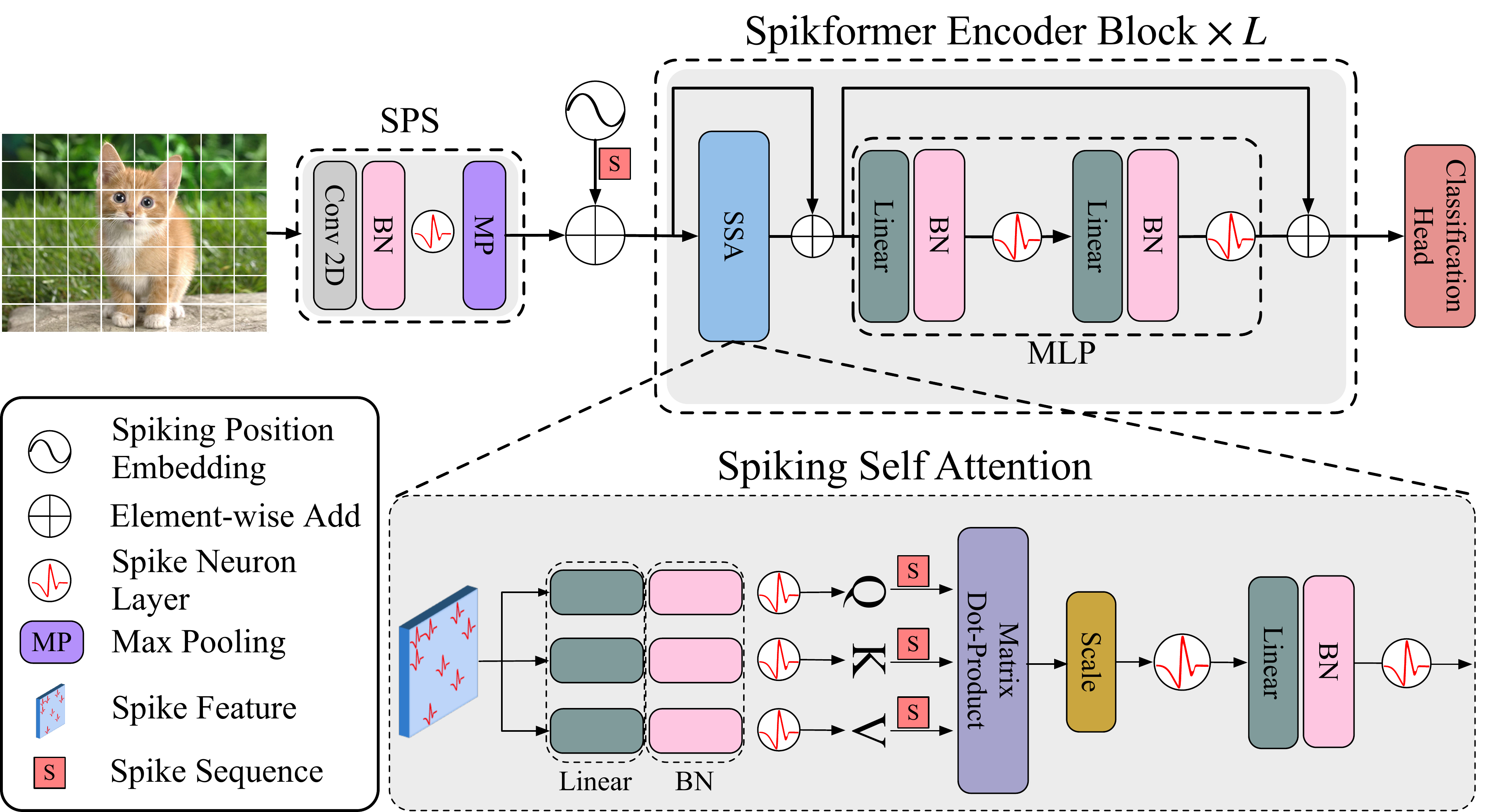}
\end{tabular}
\vspace{-8pt}
\end{center}
\caption{\small The overview of \oursfull (\oursabbrv), which consists of a spiking patch splitting module (SPS), a \oursabbrv encoder and a Linear classification head. We empircally find that the layer normalization (LN) does not apply to SNNs, so we use batch normalization (BN) instead.
}
\label{fig:model}
% \vspace{-20pt}
\vspace{-4mm}
\end{figure}}

We propose \oursfull(\oursabbrv), which incorporates the self-attention mechanism and Transformer into the spiking neural networks (SNNs) for enhanced learning capability.
Now we explain the overview and components of \oursabbrv one by one.
\vspace{-4mm}
\subsection{Overall Architecture}\label{sec:overview}
\vspace{-2mm}
An overview of \oursabbrv is depicted in Figure \ref{fig:model}. Given a 2D image sequence $I\in \mathbb{R}^{T \times C\times H\times W}$\footnote{In the neuromorphic dataset the data shape is $I \in \mathbb{R}^{T \times C\times H\times W}$, where $T$, $C$, $H$, and $ W$ denote time step, channel, height and width, respectively. A 2D image $I_s \in \mathbb{R}^{C\times H\times W}$ in static datasets need to be repeated $T$ times to form a sequence of images.}, the Spiking Patch Splitting (SPS) module linearly projects it to a $D$ dimensional spike-form feature vector and splits it into a sequence of $N$ flattened spike-form patches $x$. Float-point-form position embedding cannot be used in SNNs. We employ a conditional position embedding generator \citep{chu2021twins} to generate spike-form relative position embedding (RPE) and add the RPE to patches sequence $x$ to get $X_0$. The conditional position embedding generator contains a 2D convolution layer (Conv2d) with kernel size $3$, batch normalization (BN), and spike neuron layer ($\mathcal{SN}$). Then we pass the $X_0$ to the $L$-block \oursabbrv encoder. Similar to the standard ViT encoder block, a \oursabbrv encoder block consists of a Spiking Self Attention (SSA) and an MLP block. Residual connections are applied in both the SSA and MLP block. As the main component in \oursabbrv encoder block, \oursattn offers an efficient method to model the local-global information of images using spike-form Query ($Q$), Key ($K$), and Value ($V$) without softmax, which will be analyzed in detail in Sec. \ref{sec:ssa}. A global average-pooling (GAP) is utilized on the processed feature from \oursabbrv encoder and outputs the $D$-dimension feature which will be sent to the fully-connected-layer classification head (CH) to output the prediction $Y$.
\oursabbrv can be written as follows:
\begin{align}
&x={\rm{SPS}}\left(I\right), &&{{I}} \in \mathbb{R}^{T \times C\times H\times W}, x\in \mathbb{R}^{T\times N\times D},  \\
&{\rm{RPE}}={\mathcal{SN}}({\rm{BN}}(({\rm{Conv2d}}(x)))), &&{\rm{RPE}}\in \mathbb{R}^{T \times N\times D} \\
& X_0 = x + {\rm{RPE}}, &&X_0\in \mathbb{R}^{T \times N\times D}\\
& X^{\prime}_l = {\rm{SSA}}(X_{l-1}) + X_{l-1}, &&X^{\prime}_l\in \mathbb{R}^{T \times N\times D},l=1...L \\
& X_l = {\rm{MLP}}(X^{\prime}_l) + X^{\prime}_l, &&X_l\in \mathbb{R}^{T \times N\times D}, l=1...L \\
& Y = \op{CH}(\op{GAP}(X_L))
\end{align}

% We use  \citep{chu2021twins} to add position information to each image patch. Then the patches sequence is fed to \oursabbrv blocks. The main component in the \oursabbrv encoder block is Spiking Self-Attention (\oursattn). 

%  and  is applied to split the input image and generate spike-form patch embedding.

\subsection{Spiking Patch Splitting}\label{sec:patch_transformer}
As shown in Figure \ref{fig:model}, the Spiking Patch Splitting (SPS) module aims to linearly project an image to a $D$ dimensional spike-form feature and split the feature into patches with a fixed size. SPS can contain multiple blocks. Similar to the convolutional stem in Vision Transformer~\citep{xiao2021early,hassani2021escaping}, we apply a convolution layer in each SPS block to introduce inductive bias into \oursabbrv. Specifically, given an image sequence ${{I}} \in \mathbb{R}^{T \times C\times H\times W}$:
\begin{align}
x={\mathscr{MP}}\left({\mathcal{SN}}({\rm{BN}}(({\rm{Conv2d}}(I))))\right)
\end{align}
where the Conv2d and $\mathscr{MP}$ represent the 2D convolution layer (stride-1, $3\times 3$ kernel size) and max-pooling, respectively. The number of SPS blocks can be more than $1$. When using multiple SPS blocks, the number of output channels in these convolution layers is gradually increased and finally matches the embedding dimension of patches. For example, given an output embedding dimension~${D}$ and a four-block SPS module, the number of output channels in four convolution layers is~$D/8, D/4, D/2, D$. While the 2D-max-pooling layer is applied to down-sample the feature size after SPS block with a fixed size. After the processing of SPS, $I$ is split into an image patches sequence $x\in \mathbb{R}^{ T \times N\times D}$.

\subsection{Spiking Self Attention Mechanism}\label{sec:ssa}
\oursabbrv encoder is the main component of the whole architecture, which contains the Spiking Self Attention (SSA) mechanism and MLP block. In this section we focus on SSA, starting with a review of vanilla self-attention (VSA). Given an input feature sequence $X \in \mathbb{R}^{T \times N\times D}$, the VSA in ViT has three float-point key components, namely query ($Q_{\mathcal{F}}$), key ($K_{\mathcal{F}}$), and value ($V_{\mathcal{F}}$) which are calculated by learnable linear matrices $W_Q, W_K, W_V \in\mathbb{R}^{D\times D }$ and $X$:
\begin{align}
Q_{\mathcal{F}} = XW_Q,~ K_{\mathcal{F}}=XW_K,~ V_{\mathcal{F}} = XW_V
\end{align}
where ${\mathcal{F}}$ denotes the float-point form. The output of vanilla self-attention can be computed as:
\begin{align}\label{eq:vsa}
{\rm{VSA}}(Q_{\mathcal{F}},K_{\mathcal{F}},V_{\mathcal{F}})={\rm{Softmax}}\left(\frac{Q_{\mathcal{F}}K_{\mathcal{F}}^{\rm{T}}}{\sqrt{d}}\right)V_{\mathcal{F}}
\end{align}
where $d={D}/{H}$ is the feature dimension of one head and $H$ is the head number. Converting the float-point-form Value ($V_{\mathcal{F}}$) into spike form ($V$) can realize the direct application of VSA in SNNs, which can be expressed as:
\begin{align}\label{eq:vsainsnn}
{\rm{VSA}}(Q_{\mathcal{F}},K_{\mathcal{F}},V)={\rm{Softmax}}\left(\frac{Q_{\mathcal{F}}K_{\mathcal{F}}^{\rm{T}}}{\sqrt{d}}\right)V
\end{align}
However, the calculation of VSA is not applicable in SNNs for two reasons. 1) The float-point matrix multiplication of $Q_{\mathcal{F}}, K_{\mathcal{F}}$ and softmax function which contains exponent calculation and division operation, do not comply with the calculation rules of SNNs. 
2) The quadratic space and time complexity of the sequence length of VSA do not meet the efficient computational requirements of SNNs. 

We propose Spiking Self-Attention (\oursattn), which is more suitable for SNNs than the VSA, as shown in Figure \ref{fig:ssa}(b) and the bottom of Figure \ref{fig:model}. The query ($Q$), key ($K$), and Value ($V$) are computed through learnable matrices firstly. Then they become spiking sequences via different spike neuron layers:
\begin{align}
Q = {{\mathcal{SN}}_Q}(\op{BN}(XW_Q)), K={{\mathcal{SN}_K}}(\op{BN}(XW_K)), V = {{\mathcal{SN}_V}}(\op{BN}(XW_V))
\label{eq:spikeqkv}
\end{align}
where $Q,K,V \in \mathbb{R}^{T \times N\times D}$. We believe that the calculation process of the attention matrix should use pure spike-form Query and Key(only containing 0 and 1). Inspired by vanilla self-attention \citep{vaswani2017attention}, we add a scaling factor $s$ to control the large value of the matrix multiplication result. $s$ does not affect the property of \oursattn. As shown in Figure \ref{fig:model}, the spike-friendly \oursattn is defined as:
\begin{align}\label{eq:ssa}
&{\rm{\oursattn}}^{'}(Q,K,V)={\mathcal{SN}}\left({Q}~{K^{\rm{T}}}~V * s\right)\\
&{\rm{\oursattn}}(Q,K,V)={\mathcal{SN}}(\op{BN}(\op{Linear}({\rm{\oursattn}}^{'}(Q,K,V)))).
\end{align}
The single-head \oursattn introduced here can easily be extended to the multi-head \oursattn, which is detailed in the appendix \ref{sec:mhssa}. \oursattn is independently conducted on each time step and seeing more details in appendix \ref{sec:ssa and T}. As shown in Eq. (\ref{eq:ssa}), \oursattn cancels the use of softmax to normalize the attention matrix in Eq.~(\ref{eq:vsa}) and directly multiplies $Q,K$ and $V$. An intuitive calculation example is shown in Figure \ref{fig:ssa}(b). 
The softmax is unnecessary in our \oursattn, and it even hinders the implementation of self-attention to SNNs.
Formally,
%\oursattn already ensures the non-negativeness of the attention map $QK^{\rm{T}}$.
%As it has validated that the most key property of the softmax function in VSA is making the attention map to be non-negative~\citep{qin2022cosformer}, we argue that \oursattn also satisfies the above key point: 
based on Eq. (\ref{eq:spikeqkv}), the spike sequences $Q$ and $K$ output by the spiking neuron layer $\mathcal{SN}_Q$ and $\mathcal{SN}_k$ respectively, are naturally non-negative ($0$ or $1$), resulting in a non-negative attention map.
\oursattn only aggregates these relevant features and ignores the irrelevant information.
Hence it does not need the softmax to ensure the non-negativeness of the attention map.
%Realizing self-attention in SNNs has natural advantages. 
Moreover, compared to the float-point-form $X_{\mathcal{F}}$ and $V_{\mathcal{F}}$ in ANNs, the input $X$ and the Value $V$ of self-attention in SNNs are in spike form, containing limited information.
The vanilla self-attention (VSA) with float-point-form $Q_{\mathcal{F}}, K_{\mathcal{F}}$ and softmax is redundant for modeling the spike-form $X, V$, which cannot get more information from $X, V$ than \oursattn. 
That is, \oursattn is more suitable for SNNs than the VSA.
%quanhong: check

We conduct experiments to validate the above insights by comparing the proposed \oursattn with four different calculation methods of the attention map, as shown in Tab. \ref{tab: ablation1}. 
\begin{wraptable}[15]{r}{0.75\textwidth}
  \small
\vspace{-4mm}
  \centering
  \tabcolsep=0.25cm
  \renewcommand{\arraystretch}{0.22}
  \caption{\small Analysis of the \oursattn's rationality. We replace \oursattn with other attention variants and keep the remaining network structure in \oursabbrv unchanged. We show the accuracy (Acc) on CIFAR10-DVS \citep{li2017cifar10}, CIFAR10/100 \citep{krizhevsky2009learning}. OPs (M) is the number of operations (For $\rm{A_I},\rm{A_{LeakyReLU}},\rm{A_{ReLU}}$ and $\rm{A_{softmax}}$, OPs is FLOPs, and SOPs is ignored; For $\rm{A_{\oursattn}}$, it is SOPs.) and $\rm P$ ($\mu \rm{J}$) is the theoretical energy consumption to perform one calculation among $Q,K,V$.}
  \begin{tabular}{lccc}
  \toprule
  &  {CIFAR10-DVS} & {CIFAR10} &{CIFAR100} \\
\midrule
  &&  {Acc/OPs (M)/P ($\mu {\rm{J}}$)} & \\
%   {Acc/SOPs/P($\mu \rm{J}$)} &{Acc/OPs/P($\mu \rm{J}$)}
\midrule
$\rm{A_I}$  &79.40/16.8/{77} &  93.96/6.3/{29} & 76.94/6.3/{29} \\
\midrule
$\rm{A_{LeakyReLU}}$   &79.80/16.8/{77}	&93.85/6.3/{29} &76.73/6.3/{29}     \\
\midrule
$\rm{A_{ReLU}}$   &79.40/{16.8/77} &  94.34/{6.3/29} & 77.00/{6.3/29}       \\
\midrule
$\rm{A_{softmax}}$ & 80.00/19.1/{88} &  94.97/6.6/{30} & \textbf{77.92}/6.6/{30} \\
\midrule
$\rm{A_{\oursattn}}$  & \textbf{80.90/0.66/{0.594}} &  \textbf{95.19}/\textbf{1.1}/{\textbf{0.990}} & 
{77.86}/\textbf{1.3}/{\textbf{1.170}}\\
 \bottomrule
\end{tabular}
%\end{table*}
\label{tab: ablation1}
\end{wraptable}
$\rm{A_I}$ denotes multiplying the float-points $Q$ and $K$ directly  to get the attention map, which preserves both positive and negative correlation. 
$\rm{A_{ReLU}}$ uses the multiplication between ${\rm{ReLU}}(Q)$ and ${\rm{ReLU}}(K)$ to obtain the attention map.  $\rm{A_{ReLU}}$ retains the positive values of $Q, K$ and sets the negative values to $0$, while $\rm{A_{LeakyReLU}}$ still retains the negative points. $\rm{A_{softmax}}$ means the attention map is generated following VSA. The above four methods use the same \oursabbrv framework and weight the spike-form $V$. 
From Tab. \ref{tab: ablation1}, the superior performance of our $\rm{A_{SSA}}$ over $\rm{A_I}$ and $\rm{A_{LeakyReLU}}$ proves the superiority of ${\mathcal{SN}}$. %naturally retaining positive values.
The reason why $\rm{A_{SSA}}$ is better than $\rm{A_{ReLU}}$ may be that $\rm{A_{SSA}}$ has better non-linearity in self-attention. 
By comparing with $\rm{A_{softmax}}$, $\rm{A_{\oursattn}}$ is competitive, which even surpasses $\rm{A_{softmax}}$ on CIFAR10DVS and CIFAR10.
%quanhong: add citation
This can be attributed to \oursattn being more suitable for spike sequences ($X$ and $V$) with limited information than VSA. 
Furthermore, the number of operations and theoretical energy consumption required by the $\rm{A_{SSA}}$ to complete the calculation of $Q, K, V$ is much lower than that of the other methods.

\oursattn is specially designed for modeling spike sequences. The $Q, K$, and $V$ are all in spike form, which degrades the matrix dot-product calculation to logical AND operation and summation operation. 
We take a row of Query $q$ and a column of Key $k$ as a calculation example: $\sum_{i=1}^{d}{q_i}{k_i}=\sum_{q_i=1}k_i$.
Also, as shown in Tab. \ref{tab: ablation1}, \oursattn has a low computation burden and energy consumption due to sparse spike-form $Q, K$ and $V$ (Figure. \ref{fig:fr}) and simplified calculation.
In addition, the order of calculation between $Q, K$ and $V$ is changeable: $QK^{\rm{T}}$ first and then $V$, or $K^{\rm{T}}V$ first and then $Q$. When the sequence length $N$ is bigger than one head dimension $d$, the second calculation order above will incur less computation complexity $(O(Nd^2))$ than the first one $(O(N^2d))$.
\oursattn maintains the biological plausibility and computationally efficient properties throughout the whole calculation process.

\vspace{-4mm}
\section{Experiments}
\vspace{-4mm}
We conduct experiments on both static datasets CIFAR,
%quanhong: add citation
ImageNet \citep{deng2009imagenet}, and neuromorphic datasets CIFAR10-DVS, DVS128 Gesture \citep{amir2017dvsg} to evaluate the performance of \oursabbrv. 
The models for conducting experiments are implemented based on Pytorch \citep{paszke2019pytorch}, SpikingJelly \footnote{\url{https://github.com/fangwei123456/spikingjelly}} and Pytorch image models library (Timm) \footnote{\url{https://github.com/rwightman/pytorch-image-models}}.
We train the \oursabbrv from scratch and compare it with current SNNs models in Sec. \ref{sec:sdc} and \ref{sec:ndc}. We conduct ablation studies to show the effects of the \oursattn module and \oursabbrv in Sec. \ref{sec:ablation}.

\subsection{Static datasets classification}\label{sec:sdc}

\begin{table}[t]
\caption{Evaluation on ImageNet. Param refers to the number of parameters. Power is the average theoretical energy consumption when predicting an image from ImageNet test set, whose calculation detail is shown in Eq. \ref{eq:flop}. \oursabbrv-$L$-$D$ represents a \oursabbrv model with $L$ \oursabbrv encoder blocks and $D$ feature embedding dimensions. The train loss, test loss and test accuracy curves are shown in appendix \ref{sec:imagenet_curve}. OPs refers to SOPs in SNN and FLOPs in ANN-ViT.}
\begin{center}
\resizebox{0.9\textwidth}{!}{
\begin{tabular}{ccccccc}
\toprule
 \multicolumn{1}{c}{\bf Methods} &\multicolumn{1}{c}{\bf Architecture}
&\multicolumn{1}{c}{\bf \tabincell{c}{Param\\ (M)}}
&\multicolumn{1}{c}{\bf \tabincell{c}{OPs\\ (G)}}
&\multicolumn{1}{c}{\bf \tabincell{c}{Power\\ (mJ)}}
&\multicolumn{1}{c}{\bf\tabincell{c}{Time\\Step}}
&\multicolumn{1}{c}{\bf Acc}\\
 \midrule
  Hybrid training\citep{rathi2020enabling} &ResNet-34 &21.79 &- &- &250 &61.48\\
    \multicolumn{1}{c}{\multirow{2}{*}{TET\citep{deng2021temporal}}} 
    & Spiking-ResNet-34 &21.79 &- &- &6 & {64.79} \\
    & SEW-ResNet-34 &21.79 &- &- &4 & {68.00} \\
    \multicolumn{1}{c}{\multirow{2}{*}{Spiking ResNet\citep{hu2018residual}}} 
    &ResNet-34 &21.79 &65.28 & {59.295} & 350 & 71.61 \\
    &ResNet-50 &25.56&78.29 & {70.934} & 350 & 72.75 \\
    \multicolumn{1}{c}{\multirow{1}{*}{STBP-tdBN\citep{zheng2021going}}} &Spiking-ResNet-34 &21.79 &6.50 &{6.393} & 6 & 63.72 \\
    % &Spiking-ResNet-50 &25.56 & & 6 & 63.72 \\
    \multirow{4}{*}{SEW ResNet\citep{fang2021deep}}  
    & SEW-ResNet-34 &21.79 &3.88 &{4.035} &4 & 67.04 \\
    & SEW-ResNet-50 &25.56  &4.83 &{4.890} &4 & 67.78 \\
    & SEW-ResNet-101 &44.55  &9.30 &{8.913} &4 & 68.76 \\
    & SEW-ResNet-152 &60.19  &13.72 &{12.891} &4 & 69.26 \\
 \midrule
  {Transformer}   & {Transformer-8-512} & {29.68} & {8.33} & {38.340} & {1} & {\textbf{80.80}} \\

 \midrule
    \multicolumn{1}{c}{\multirow{5}{*}{\textbf{\oursabbrv}}}  &\multicolumn{1}{c}{\multirow{1}{*}{\oursabbrv-8-384}} &16.81&6.82 &{7.734} &4 & 70.24\\
    &\multicolumn{1}{c}{\multirow{1}{*}{\oursabbrv-6-512}}&23.37&8.69 &{9.417} & 4 & 72.46\\
    &\multicolumn{1}{c}{\multirow{1}{*}{\oursabbrv-8-512}}&29.68&11.09&{11.577} & 4 & \textbf{73.38}\\
    &\multicolumn{1}{c}{\multirow{1}{*}{\oursabbrv-10-512}}&36.01&13.67 &{13.899} & 4 & \textbf{73.68}\\
    % &\multicolumn{1}{c}{\multirow{1}{*}{\oursabbrv-12-512}}&42.35 & & & 4 & 72.50\\
    &\multicolumn{1}{c}{\multirow{1}{*}{\oursabbrv-8-768}} &66.34&22.09 &{21.477} & 4 & \textbf{74.81}\\
    \hline
\end{tabular}}
\end{center}
\label{tab:imagenet}
\vspace{-6mm}
\end{table}

\textbf{ImageNet} contains around $1.3$ million $1,000$-class images for training and $50,000$ images for validation. The input size of our model on ImageNet is set to the default $224\times 224$. 
The optimizer is AdamW and the batch size is set to $128$ or $256$ during $310$ training epochs with a cosine-decay learning rate whose initial value is $0.0005$. The scaling factor is $0.125$ when training on ImageNet and CIFAR.
A four-block SPS splits the image into $196$ $16 \times 16$ patches. Following~\citep{yuan2021tokens}, standard data augmentation methods, such as random augmentation, mixup, and cutmix, are also used in training.
\begin{wrapfigure}{r}{0.25\textwidth}
\centering
\includegraphics[height=3in,trim={0in 0in 0.15in 0in},clip]{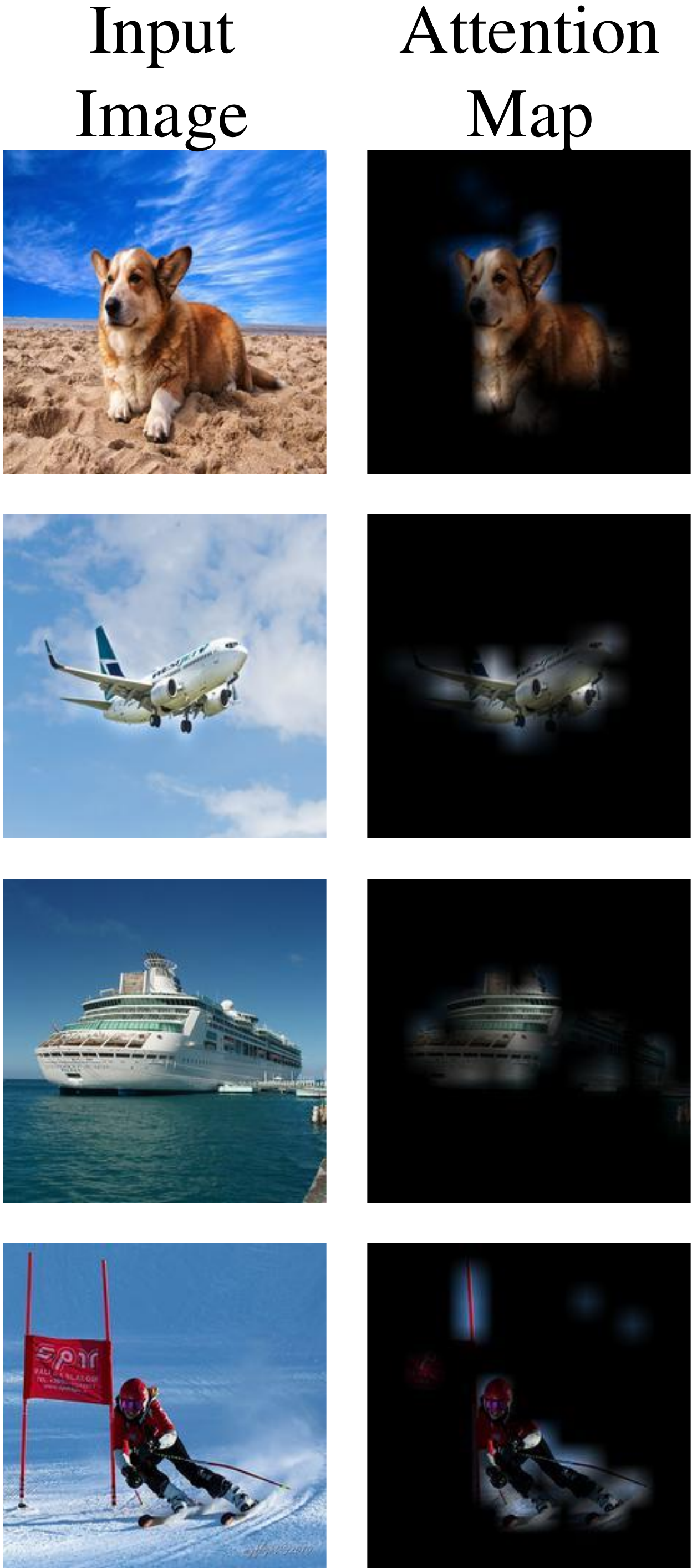}
\caption{\small Attention map examples of \oursattn. The black region is 0.}
\label{fig:attention_map_examples}
\vspace{-4mm}
\end{wrapfigure}We try a variety of models with different embedding dimensions and numbers of transformer blocks for ImageNet, which has been shown in Tab. \ref{tab:imagenet}.
We also give a comparison of synaptic operations (SOPs) \citep{merolla2014million} and theoretical energy consumption.
From the results, it can be seen that our \oursabbrv achieves a significant accuracy boost on the ImageNet compared with the current best SNNs models. 
% To our knowledge, this is the first directly trained SNNs work that achieves $74\%$ using 4 time steps on ImageNet. 
In particular, our comparison first starts from our smallest model with other models. 
The \oursabbrv-$8$-$384$
with $16.81$M parameters has $70.24\%$ top-1 accuracy when trained from scratch on ImageNet, which outperforms the best the current best direct-train model SEW-ResNet-152: $69.26\%$ with $60.19$M.
In addition, the SOPs and the theoretical energy consumption of \oursabbrv-$8$-$384$ (6.82G, {7.734mJ}) are lower compared with the SEW-ResNet-152 (13.72G, {12.891mJ}). The $29.68$M model \oursabbrv-$8$-$512$ has already achieved state-of-the-art performance with $73.38\%$, which is even higher than the converted model \citep{hu2018residual} ($72.75\%$) using $350$ time steps.
As the number of \oursabbrv blocks increases, the classification accuracy of our model on ImageNet is also getting higher. The \oursabbrv-$10$-$512$ obtains $73.68\%$ with $42.35$M.
The same happens when gradually increasing the embedding dimension, where \oursabbrv-$8$-$768$ further improves the performance to $74.81\%$ and significantly outperforms the SEW-ResNet-152 model by $5.55\%$. {ANN-ViT-8-512 is $7.42\%$ higher than Spikformer-8-512, but the theoretical energy consumption is $3.31 \times$ of \oursabbrv-8-512.} In Figure \ref{fig:attention_map_examples}, we show the attention map examples of the last encoder block in \oursabbrv-$8$-$512$ at the fourth time step. \oursattn can capture image regions associated with classification semantics and set irrelevant regions to 0 (black region), and is shown to be effective, event-driven, and energy-efficient.

\textbf{CIFAR} provides $50,000$ train and $10,000$ test images with $32\times 32$ resolution. The batch size is set to $128$. A four-block SPS (the first two blocks do not contain the max-pooling layer) splits the image into $64$ $4\times 4$ patches.  Tab. \ref{tab:sd} shows the accuracy of \oursabbrv compared with other models on CIFAR.
As shown in Tab. \ref{tab:sd}, \oursabbrv-$4$-$384$ achieves $95.19\%$ accuracy on CIFAR10, which is better than the TET ($94.44\%$) and ResNet-19 ANN ($94.97\%$). The performance is improved as the dimensions or blocks increase. 
Specifically, \oursabbrv-4-384 improves by $1.25\%$ compared to \oursabbrv-4-256 and improves by $0.39\%$ compared to \oursabbrv-2-384. We also find that extending the number of training epochs to 400 can improve the performance (\oursabbrv-4-384 400E achieves $0.32\%$ and $0.35\%$ advance compared to \oursabbrv-4-384 on CIFAR10 and CIFAR100). The improvement of the proposed \oursabbrv on complex datasets such as CIFAR100 is even higher. \oursabbrv-4-384 ($77.86\%, 9.32 \rm M$) obtains a significant improvement of $2.51\%$ compared with ResNet-19 ANN ($75.35\%, 12.63\rm M$) model.
{The ANN-Transformer model is $1.54\%$ and $3.16\%$ higher than Spikformer-4-384, respectively. As shown in appendix \ref{sec:transfer_learning}, transfer learning can achieve higher performance on CIFAR based on pre-trained \oursabbrv, which demonstrates high transfer ability.} 
%ResNet-20\textbf{/}
%250\textbf{/}
%0.27\textbf{/}

\begin{table}%[t]
\small
\caption{Performance comparison of our method with existing methods on CIFAR10/100. Our method improves network performance across all tasks. * denotes self-implementation results by \cite{deng2021temporal}. Note that Hybrid training \citep{rathi2020enabling} adopts ResNet-20 for CIFAR10 and VGG-11 for CIFAR100.}
\begin{center}
\resizebox{0.9\textwidth}{!}{
\begin{tabular}{ccccccc}
\toprule
  \multicolumn{1}{c}{\bf Methods} &\multicolumn{1}{c}{\bf Architecture} &\multicolumn{1}{c}{\bf \tabincell{c}{Param\\ (M)}}
&\bf\tabincell{c}{Time\\Step} &\bf\tabincell{c}{CIFAR10\\Acc} &\bf\tabincell{c}{CIFAR100\\Acc}\\
\midrule
    Hybrid training\citep{rathi2020enabling} &VGG-11 &9.27 &125 &92.22 &67.87\\
    Diet-SNN\citep{rathi2020diet} &ResNet-20 &0.27 &10\textbf{/}5  & 92.54& 64.07\\
    STBP\citep{wu2018spatio} &CIFARNet &17.54&12 & 89.83&-\\
    STBP NeuNorm\citep{wu2019direct} &CIFARNet &17.54 &12 &90.53& -\\
    TSSL-BP\citep{zhang2020temporal} &CIFARNet &17.54 &5 &91.41& -\\
    \multicolumn{1}{c}{\multirow{1}{*}{{STBP-tdBN\citep{zheng2021going}}}} &\multicolumn{1}{c}{\multirow{1}{*}{ResNet-19}} &12.63 & 4 & 92.92 & 70.86\\
    TET\citep{deng2021temporal}  &\multicolumn{1}{c}{\multirow{1}{*}{ResNet-19}} &12.63 & 4 & \textbf{94.44}& \textbf{74.47}\\
\midrule
\multicolumn{1}{c}{\multirow{2}{*}{\textbf{ANN}}} 
     &ResNet-19* &12.63 &1 &\textbf{94.97} & \textbf{75.35}\\
     & {Transformer-4-384} & {9.32} &1 &{\textbf{96.73}} & {\textbf{81.02}} \\
\midrule
    \multicolumn{1}{c}{\multirow{4}{*}{\textbf{\oursabbrv}}}  &\multicolumn{1}{c}{\multirow{1}{*}{\oursabbrv-4-256}}&4.15 & 4 & 93.94 & \textbf{75.96}\\
    &\multicolumn{1}{c}{\multirow{1}{*}{\oursabbrv-2-384}}&5.76 & 4 & \textbf{94.80} & \textbf{76.95}\\
    &\multicolumn{1}{c}{\multirow{1}{*}{\oursabbrv-4-384}}&9.32 & 4 & \textbf{95.19} & \textbf{77.86}\\
    &\multicolumn{1}{c}{\multirow{1}{*}{\oursabbrv-4-384 400E}}&9.32 & 4 & \textbf{95.51} & \textbf{78.21}\\

\bottomrule
\end{tabular}}
\end{center}
\label{tab:sd}
\vspace{-6mm}
\end{table}

\vspace{-4mm}
\subsection{Neuromorphic datasets classification}\label{sec:ndc}
DVS128 Gesture is a gesture recognition dataset that contains 11 hand gesture categories from 29 individuals under 3 illumination conditions. CIFAR10-DVS is also a neuromorphic dataset converted from the static image dataset by shifting image samples to be captured by the DVS camera, which provides $9,000$ training samples and $1,000$ test samples.

%The implementation details are shown as follows. 
For the above two datasets of image size $128 \times 128$, we adopt a four-block SPS. The patch embedding dimension is $256$ and the patch size is $16 \times 16$. We use a shallow \oursabbrv with $2$ transformer encoder blocks. The \oursattn contains $8$ and $16$ heads for DVS128 Gesture and CIFAR10-DVS, respectively. The time-step of the spiking neuron is $10$ or $16$. The training epoch is $200$ for DVS128 Gesture and $106$ for CIFAR10-DVS. The optimizer is AdamW and the batch size is set to $16$. The learning rate is initialized to $0.1$ and reduced with cosine decay. We apply data augmentation on CIFAR10-DVS according to \citep{li2022neuromorphic}. We use a learnable parameter as the scaling factor to control the $QK^{\rm{T}}V$ result.
% For transformer block, all linear projection operations are replaced with 1D-convolution operations. 

The classification performance of \oursabbrv as well as the compared state-of-the-art models on neuromorphic datasets is shown in Tab. \ref{tab:nd}. It can be seen that our model achieves good performance on both datasets by using a $2.59$M model. On DVS128 Gesture, we obtain an accuracy of $98.2\%$ with 16-time steps, which is higher than SEW-ResNet ($97.9\%$). Our result is also competitive compared with TA-SNN ($98.6\%$, $60$ time steps) \citep{yao2021temporal} which uses floating-point spikes in the forward propagation.
% except the models that achieve improved performance by replacing binary spikes with  .
On CIFAR10-DVS, we achieve a $1.6\%$ and $3.6\%$ better accuracy than the SOTA methods DSR ($77.3\%$) with binary spikes using 10 steps and 16 steps respectively. TET is not an architecture-based but a loss-based method which achieves $83.2\%$ using long epochs ($300$) and $9.27$M VGGSNN, so we do not compare with it in the table. 
\begin{table}[t]
\footnotesize
  \centering
  \caption{Performance comparison to the state-of-the-art (SOTA) methods on two neuromorphic datasets. Bold font means the best; $^*$ denotes with Data Augmentation.}
  \begin{adjustbox}{max width=\linewidth} 
  \begin{tabular}{lccccccc}
  \toprule 
  \multirow{3}*{{Method}} &
  \multirow{3}*{{Spikes}} & \multicolumn{2}{c}{{CIFAR10-DVS}} &  \multicolumn{2}{c}{{DVS128}} \\
  \cmidrule(l{2pt}r{2pt}){3-4}\cmidrule(l{2pt}r{2pt}){5-6}\cmidrule(l{2pt}r{2pt}){7-8}
    &{} & {$T$ Step} & {Acc} & {$T$ Step} & {Acc} \\
\midrule
  LIAF-Net~\citep{wu2021liaf}\textsuperscript{TNNLS-2021} & \xmark& 10 & 70.4 & 60 & 97.6 \\ 
    TA-SNN~\citep{yao2021temporal}\textsuperscript{ICCV-2021} & \xmark& 10 & 72.0 &60 & 98.6 \\ 
  \midrule
%   SLAYER~\citep{shrestha2018slayer}\textsuperscript{NeurIPS-2018} & \cmark & - & - & 1600 & 93.4 \\
%   HATS~\citep{sironi2018hats}\textsuperscript{CVPR-2018} & N/A & N/A & 52.4 &  - & - \\ 
%   DART~\citep{ramesh2019dart}\textsuperscript{TPAMI-2019} & N/A & N/A & 65.8 &  - & - \\ 
%   NeuNorm~\citep{wu2019direct}\textsuperscript{AAAI-2019} & \cmark & 230-292 & 60.5 &  - & - \\ 
  Rollout~\citep{kugele2020efficient}\textsuperscript{Front. Neurosci-2020} & \cmark & 48 & 66.8 & 240 & 97.2 \\ 
  DECOLLE~\citep{kaiser2020synaptic}\textsuperscript{Front. Neurosci-2020} & \cmark & - & - & 500 & 95.5 \\ 

  tdBN~\citep{zheng2021going}\textsuperscript{AAAI-2021} & \cmark& 10 & 67.8 &40 & 96.9 \\ 
  PLIF~\citep{fang2021incorporating}\textsuperscript{ICCV-2021} & \cmark& 20 & 74.8 &20 & 97.6 \\ 

  SEW-ResNet~\citep{fang2021deep}\textsuperscript{NeurIPS-2021} & \cmark& 16 & 74.4 &16 & 97.9 \\ 
  Dspike~\citep{li2021differentiable}\textsuperscript{NeurIPS-2021} & \cmark& 10 & 75.4$^*$ & - & - \\ 
  SALT~\citep{kim2021optimizing}\textsuperscript{Neural Netw-2021} & \cmark & 20 & 67.1 & - & - \\ 
%   TET~\citep{deng2021temporal}\textsuperscript{ICLR-2022} &\cmark & 10 & 83.2$^*$ &  - & - \\ 
  DSR~\citep{meng2022training}\textsuperscript{CVPR-2022} &\cmark & 10 & 77.3$^*$ & - & - \\ 
  \midrule
  {\multirow{2}{*}{\textbf{\oursabbrv}}} & \cmark & 10 & 78.9$^*$ & 10 &96.9 \\ 
  &\cmark & 16 & \textbf{80.9}$^*$ & 16 & \textbf{98.3} \\ 
  \bottomrule
  \end{tabular}
\end{adjustbox}
   \vspace{-4mm}
\label{tab:nd}
\end{table}
\subsection{Ablation Study}\label{sec:ablation}

\textbf{Time step}
The accuracy regarding different simulation time steps of the spike neuron is shown in Tab.~\ref{tab: ablation_stu}. When the time step is $1$, our method is $1.87\%$ lower than the network with $T=4$ on CIFAR10. \oursabbrv-$8$-$512$ with $1$ time step still achieves $70.14\%$. The above results show \oursabbrv is robust under low latency (fewer time steps) conditions.

\textbf{\oursattn}
We conduct ablation studies on \oursattn to further identify its advantage.
We first test its effect by replacing \oursattn with standard vanilla self-attention. We test two cases where Value is in floating point form (\oursabbrv-$L$-$D_{w\ \rm{VSA}~V_{\mathcal{F}}}$) and in spike form (\oursabbrv-$L$-$D_{w\ \rm{VSA}}$).
\begin{wraptable}[16]{r}{0.60\textwidth}
\vspace{-4mm}
\small
  \centering
  \tabcolsep=0.1cm
  \renewcommand{\arraystretch}{0.32}
  \caption{\small Ablation study results on \oursattn, and time step.
  }
\fontsize{8pt}{12pt}\selectfont
% \tiny
% \caption{}
\begin{tabular}{llcc}
 \toprule
 Datasets& Models &\tabincell{c}{Time\\ Step} &\tabincell{c}{Top1-Acc\\(\%)}  \\
 \midrule
%   \multicolumn{1}{c}{\multirow{2}{*}{CIFAR10DVS}} & \oursabbrv-2-256$_{w\ RCV}$& 16 & 80.8 \\
%   & \oursabbrv-2-256 & 16 & 80.6 \\
%   \midrule
 \multicolumn{1}{c}{\multirow{6}{*}{CIFAR10/100}}
%  & \oursabbrv-4-384$_{w\ RCV}$ &4 &95.20/77.89 \\
%  \cmidrule(lr){2-4}
 &\multicolumn{1}{l}{\multirow{6}{*}{\oursabbrv-4-384$_{w\ \rm{SSA}}$}}
 &1 & 93.51/74.36\\
 && 2 & 93.59/76.28\\
 && 4 & 95.19/77.86\\
 && 6 & 95.34/78.61\\
   \cmidrule(lr){2-4}
&\oursabbrv-4-384$_{w\ \rm{VSA}}$&4 &94.97/77.92    \\
  &\oursabbrv-4-384$_{w\ \rm{VSA}~V_{\mathcal{F}}}$&4 &95.17/78.37    \\
%  & T2T-ViT$_{t}$-14 &81.7 \color{blue}(+2.2) &21.5 & 6.1\\
%  & T2T-ViT$_{c}$-14 &80.8 \color{blue}(+1.3) &21.3 & 4.6\\
  \midrule
 \multirow{15}{*}{\tabincell{c}{ImageNet}}
  &\oursabbrv-8-512$_{w\ \rm{I}}$&4 &\xmark   \\
 &\oursabbrv-8-512$_{w\ \rm{ReLU}}$&4 &\xmark   \\
 &\oursabbrv-8-512$_{w\ \rm{LeakyReLU}}$&4 &\xmark   \\
 &\oursabbrv-8-512$_{w\ \rm{VSA}}$&4 &72.70    \\
  &\oursabbrv-8-512$_{w\ \rm{VSA}~V_{\mathcal{F}}}$&4 &73.96    \\
%  & \oursabbrv-8-512$_{w\ RCV}$      &4 &72.70 \\
  \cmidrule(lr){2-4}
  &\multicolumn{1}{l}{\multirow{4}{*}{\oursabbrv-8-512$_{w\ \rm{SSA}}$}}
 &1 &70.14    \\
 &&2 &71.09    \\
 &&4 &73.38   \\
 &&6 &73.70  \\
\bottomrule
\end{tabular}
\label{tab: ablation_stu}
\end{wraptable}
We also test the different attention variants on ImageNet following Tab. \ref{tab: ablation1}. On CIFAR10, the performance of \oursabbrv with \oursattn is competitive compared to \oursabbrv-$4$-$384_{w\ \rm{VSA}}$ and even \oursabbrv-$4$-$384_{w\ \rm{VSA}~V_{\mathcal{F}}}$. 
On ImageNet, our \oursabbrv-8-512$_{w\ \rm{SSA}}$ outperforms \oursabbrv-$8$-$512_{w\ \rm{VSA}}$ by $0.68\%$. On CIFAR100 and ImageNet, the accuracy of \oursabbrv-$L$-$D_{w\ \rm{VSA}~V_{\mathcal{F}}}$ is better than \oursabbrv because of the float-point-form Value. The reason why the \oursabbrv-8-512$_{w\ \rm{I}}$, \oursabbrv-8-512$_{w\ \rm{ReLU}}$, and \oursabbrv-8-512$_{w\ \rm{LeakyReLU}}$ do not converge is that the value of dot-product value of Query, Key, and Value is large, which makes the surrogate gradient of the output spike neuron layer disappear. More details are in the appendix \ref{sec:converge}. In comparison, the dot-product value of the designed \oursattn is in a controllable range, which is determined by the sparse spike-form $Q$, $K$ and $V$, and makes ${\rm{\oursabbrv}}_{w\ \rm{SSA}}$ easy to converge.

% Besides, the recognition accuracy under different time step of spike neuron is also shown in Tab. \ref{tab: ablation_stu}. When the simulation time step of  spike nueron is $2$, our method is only $1.61\%$ lower than the network with $T=4$ on CIFAR10, which proves that \oursabbrv is robust under low latency conditions. The results on complex datasets such as CIFAR100 and ImageNet also verify the above advantage. Classification performance also gets better with increasing simulation step size.

%The rest of the training hyper-parameters are consistent with the CIFAR models.

\section{Conclusion}

In this work we explored the feasibility of implementing the self-attention mechanism and Transformer in Spiking Neuron Networks and propose \oursabbrv based on a new Spiking Self-Attention (\oursattn). 
Unlike the vanilla self-attention mechanism in ANNs, \oursattn is specifically designed for SNNs and spike data.
%, resulting in its biological plausibility and event-driven property.
We drop the complex operation of softmax in \oursattn, and instead perform matrix dot-product directly on spike-form Query, Key, and Value, which is efficient and avoids multiplications. 
In addition, this simple self-attention mechanism makes \oursabbrv work surprisingly well on both static and neuromorphic datasets. 
With directly training from scratch, \oursfull outperforms the state-of-the-art SNNs models. 
We hope our investigations pave the way for further research on transformer-based SNNs models.

% Though such initial performance are encouraging, challenges remain.
% One is to apply \oursabbrv to other computer vision tasks, such as detection and segmentation.
% Our results, coupled with those in \citet{carion20-detr}, indicate the promise of this approach.
% Another challenge is to continue exploring self-supervised pre-training methods.
% Our initial experiments show improvement from self-supervised pre-training, but there is still large gap between self-supervised and large-scale supervised pre-training.
% Finally, further scaling of \oursabbrv{} would likely lead to improved performance.
\clearpage
\section*{Reproducibility Statement}
%All experimental results in this paper are reproducible, and the proposed model is also supported by the complete code project. 

Our codes are based on SpikingJelly\citep{SpikingJelly}, an open-source SNN framework, and Pytorch image models library (Timm)\citep{rw2019timm}.
The experimental results in this paper are reproducible. %And the same random seed was used in the experiments to maximize reproducibility.
We explain the details of model training and dataset augmentation in the main text and supplement it in the appendix. Our codes of Spikformer models are uploaded as supplementary material and will be available on GitHub after review.

% \section*{Acknowledgements}
% The work was performed in Berlin, Z\"urich, and Amsterdam. We thank many colleagues at Google for their help, in particular Andreas Steiner for crucial help with the infrastructure and the open-source release of the code; Joan Puigcerver and Maxim Neumann for help with the large-scale training infrastructure; Dmitry Lepikhin, Aravindh Mahendran, Daniel Keysers, Mario Lu\v{c}i\'{c}, Noam Shazeer, Ashish Vaswani, and Colin Raffel for useful discussions.

\bibliography{iclr2023_conference}
\bibliographystyle{iclr2023_conference}

\newpage
\appendix
\section*{Appendix}

\section{Multihead Spiking Self Attention}
\label{sec:mhssa}
In practice, we reshape the $Q, K, V \in \mathbb{R}^{T \times N\times D}$ into multi-head form $\mathbb{R}^{T \times H \times N \times d}$, where $D = H \times d$. Then we split $Q,K,V$ into $H$ parts and run $H$ \oursattn operations, in parallel, which are called $H$-head \oursattn.
The Multihead Spiking Self Attention (MSSA) is shown in follows:
\begin{align}
   &Q = (q_1,q_2,\cdots,q_H),K = (k_1,k_2,\cdots,k_H),V = (v_1,v_2,\cdots,v_H)
   &&q,k,v \in  \mathbb{R}^{T \times N \times d}
\end{align}
{
\begin{align}
    \op{MSSA}^{'}(Q,K,V) &= [\op{SSA}^{'}_1(q_1,k_1,v_1); \op{SSA}^{'}_2(q_2,k_2,v_2); \cdots ; \op{SSA}^{'}_h(q_H,k_H,v_H)]
\end{align}
\begin{align}
    \op{MSSA}(Q,K,V)={\mathcal{SN}}(\op{BN}(\op{Linear}(\op{MSSA}^{'}(Q,K,V))))\label{eq:mssa}
\end{align}
}
\section{Spiking Self Attention and Time Step}\label{sec:ssa and T}
In practice,  $T$ is a independent dimension for spike neuron layer. In other layers, it is merged with the batch size.
\section{Experiment details}
\subsection{Training}\label{sec:training}
Unlike the standard ViT, Dropout and Droppath are not applied in \oursabbrv. We remove the layer norm before each self-attention and MLP block, and add batch norm after each linear layer instead. In all \oursabbrv models, the hidden dimension of MLP blocks is $4 \times D$, where $D$ is the embedding dimension. As in Eq. (\ref{eq:sigmoid}), we select the Sigmoid function as the surrogate function with $\alpha=4$.
\begin{align}
    \op{Sigmoid}(x)=\frac{1}{1 + \exp{(-\alpha x)}}  \label{eq:sigmoid}
\end{align}
For DVS128 Gesture, we place a 1D max-pooling layer after $Q$ and $K$ to increase the density of the data, which improves the accuracy from $97.9\%$ to $98.3\%$ in 16 time steps. We set the threshold voltage $V_{th}$ of the spike neuron layer after $QK^{\op{T}}V * s$ to $0.5$, while the others are set to $1$. 
\subsection{Theoretical synaptic operation and energy consumption calculation}\label{sec:energy}
The calculation of theoretical energy consumption requires first calculating the synaptic operations:
\begin{align}
    \op{SOPs}(l)= fr \times T \times \op{FLOPs}(l)  \label{eq:sop}
\end{align}
where $l$ is a block/layer in \oursabbrv, $fr$ is the firing rate of the input spike train of the block/layer and $T$ is the simulation time step of spike neuron. $\op{FLOPs}(l)$ refers to floating point operations of $l$, which is the number of multiply-and-accumulate (MAC) operations. And $\op{SOPs}$ is the number of spike-based accumulate (AC) operations. We estimate the theoretical energy consumption of \oursabbrv according to \citep{kundu2021hire,hu2021advancing,horowitz20141,kundu2021spike,yin2021accurate,panda2020toward,yao2022attention}. We assume that the MAC and AC operations are implemented on the 45nm hardware [12], where $E_{MAC}=4.6pJ$ and $E_{AC}=0.9pJ$. The theoretical energy consumption of \oursabbrv is calculated:

{\begin{align}
    E_{Spikformer} 
     &= E_{MAC}  \times {\rm FL}^1_{{\rm SNN~Conv}} \nonumber \\  
     &+ E_{AC} \times \left(\sum_{n=2}^{N}{\rm SOP}^n_{{\rm SNN~Conv}} + \sum_{m=1}^{M}{\rm SOP}^m_{{\rm SNN~FC}} + \sum_{l=1}^L{\rm SOP}^l_{{\rm SSA}}\right) \label{eq:flop}
\end{align}
where ${\rm FL}^1_{SNN~Conv}$ is the first layer to encode static RGB images into spike-form. Then the SOPs of $m$ SNN Conv layers, $n$ SNN Fully Connected Layer (FC) and $l$ SSA are added together and multiplied by $E_{AC}$. For ANNs, the theoretical energy consumption of block $b$ is calculated:
\begin{align}
    \op{Power}(b)= 4.6{pJ} \times \op{FLOPs}(b)  
    \label{eq:sop2}
\end{align}
For SNNs, $\op{Power}(b)$  is:
\begin{align}
    \op{Power}(b)= 0.9{pJ} \times \op{SOPs}(b)  
    \label{eq:sop3}
\end{align}}
\begin{figure}[h]
\begin{center}
\begin{tabular}{c}
\includegraphics[width=.35\textwidth]{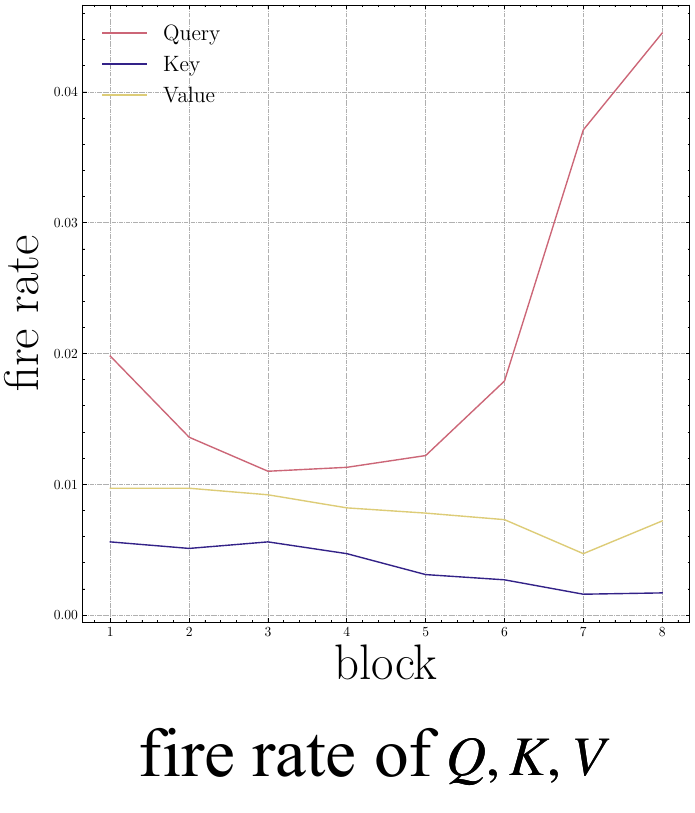}
\end{tabular}
\vspace{-8pt}
\end{center}
\caption{Fire rate of Query, Key and Value of blocks in Spikformer-$8$-$512$ on ImageNet test set.
}
\label{fig:fr}
\end{figure}
\begin{figure}[h]
\begin{center}
\begin{tabular}{c}
\includegraphics[width=.98\textwidth]{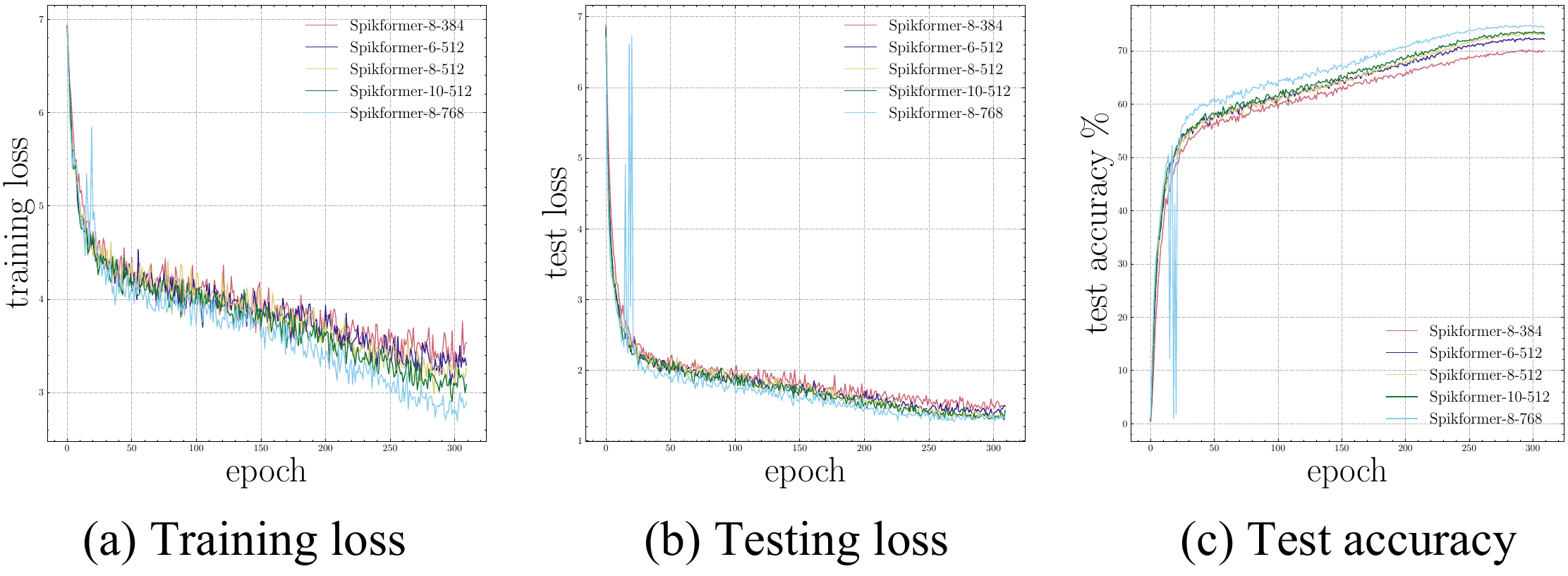}
\end{tabular}
\vspace{-8pt}
\end{center}
\caption{Training loss, testing loss and test accuracy on ImageNet.
}
\label{fig:imagenet_curve}
\end{figure}
\vspace{-2mm}
\section{Additional Results}
\subsection{Fire rate of Query, Key and Value}\label{sec:fr}
As shown in \ref{fig:fr}, the Query, Key and Value are very spare in \oursattn, causing sparse computation of \oursattn.

\subsection{Loss and Accuracy on ImageNet}\label{sec:imagenet_curve}

We show the training loss, testing loss and test accuracy of \oursabbrv in Figue. \ref{fig:imagenet_curve}. Both training and testing losses decrease as the number of \oursabbrv blocks increases or the embedding dimension increases.
\begin{table}[t]
\caption{Additional result on CIFAR10/100. \oursabbrv-4-384$_{w\ \rm{IF}}$ uses the Integrate-and-Fire neuron.}
\begin{center}
\fontsize{8pt}{12pt}\selectfont
\begin{tabular}{lcc}
 \toprule
 Models &\tabincell{c}{Time\\ Step} &\tabincell{c}{Top1-Acc\\(\%)}  \\
 \midrule
  \oursabbrv-4-384$_{w\ \rm{I}}$&1 &92.39/74.28   \\
 \oursabbrv-4-384$_{w\ \rm{ReLU}}$&1 &92.98/74.32   \\
 \oursabbrv-4-384$_{w\ \rm{LeakyReLU}}$&1 &92.88/74.31 \\
 \oursabbrv-4-384$_{w\ \rm{VSA}}$&1 &93.11/74.37   \\
  \oursabbrv-4-384$_{w\ \rm{IF}}$&4 &95.33/78.14   \\
%   \oursabbrv-4-384$_{w\ \rm{VSA}~V_{\mathcal{F}}}$&1 &/74.95    \\
\bottomrule
\end{tabular}
\end{center}
\label{tab:addition_cifar}
\vspace{-4mm}
\end{table}
\subsection{Additional Accuracy Results on CIFAR}\label{sec:cifar_ad_results}
We conduct additional experiments on CIFAR as shown in Tab. \ref{tab:addition_cifar}.

\subsection{Analysis of self-attention variants not converging on ImageNet}\label{sec:converge}
\begin{figure}[ht]
\begin{center}
\begin{tabular}{c}
\includegraphics[width=.62\textwidth]{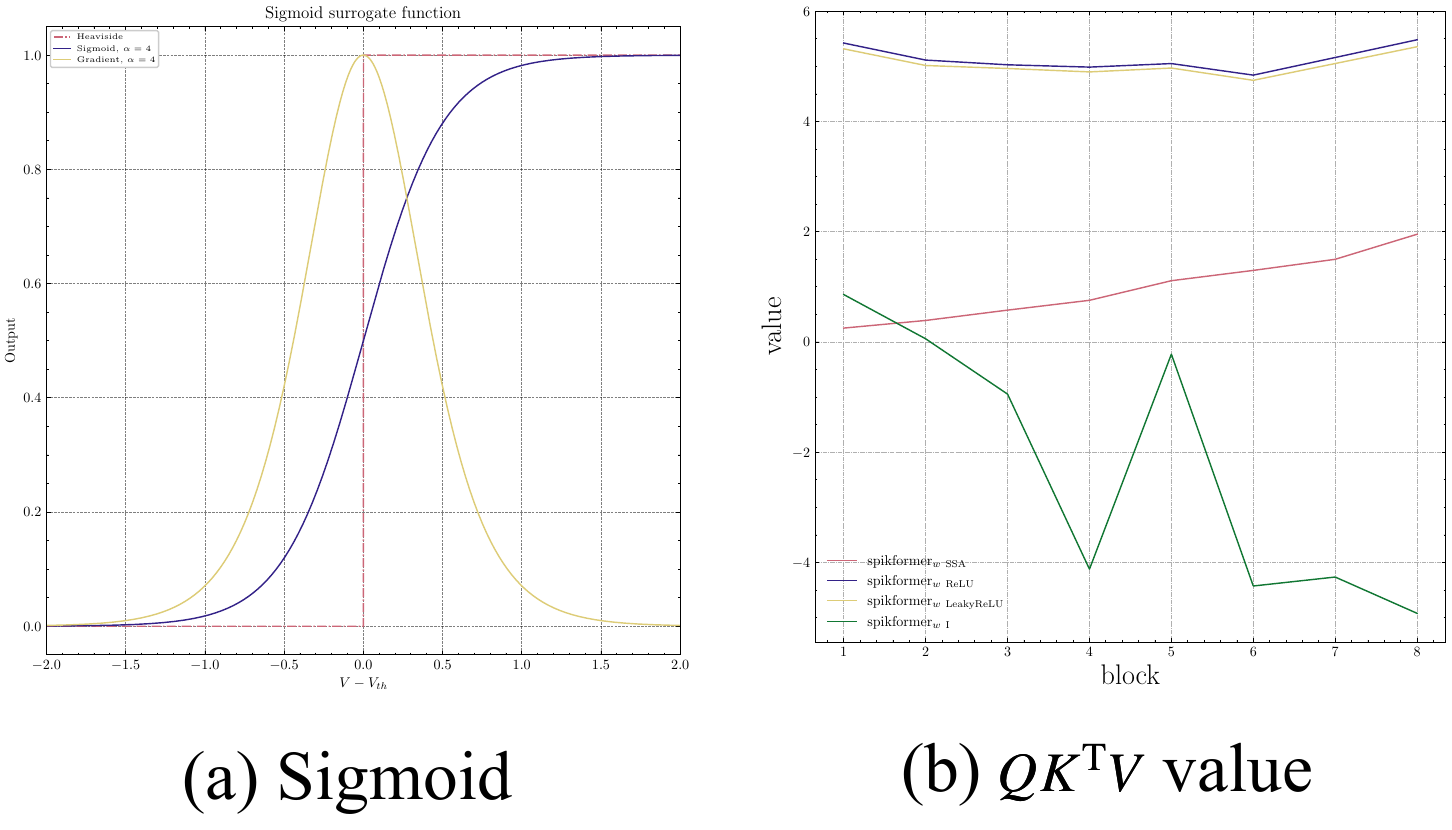}
\end{tabular}
\vspace{-8pt}
\end{center}
\caption{(a) the sigmoid surrogate function and its gradient curve. (b) the value of $QK^{\op{T}}V$.
}
\label{fig:surrogate}
\end{figure}

The reason that the three models do not converge in Tab. \ref{tab: ablation_stu} is explain as follows. As shown in Figure. \ref{fig:surrogate} (a), the gradient of sigmoid surrogate function vanishes when the difference between the average input value $V_i$ and the firing threshold $V_{th}$ is too large or too small. We collect the output value of $QK^{\op{T}}V * s$ after one training eopch of \oursabbrv-8-512$_{w\ \rm{I}}$, \oursabbrv-8-512$_{w\ \rm{ReLU}}$, \oursabbrv-8-512$_{w\ \rm{LeakyReLU}}$, and \oursabbrv-8-512$_{w\ \rm{SSA}}$, which will be sent to the spike neuron layer as the input value $V_i$, as shown in Eq. (\ref{eq:ssa}). 
Compared to the other three variants, as shown in Figure. \ref{fig:surrogate} (b), the value of $QK^{\op{T}}V * s$ in \oursabbrv-8-512$_{w\ \rm{SSA}}$ is controlled in a suitable range. Therefore, \oursattn has stable surrogate gradients during training and converges easily.

\subsection{Transfer Learning}\label{sec:transfer_learning}
{We transfer \oursabbrv to the downstream CIFAR dataset. The pre-trained \oursabbrv-4-384 and \oursabbrv-8-384/512 on ImageNet are finetuned with 60 epochs. The input size of CIFAR is $224\times224$. The remaining hyperparameters are the same as the ones directly trained on CIFAR. As shown in Tab. 7, \oursabbrv shows high transfer ability.}
\begin{table}[t]
\caption{Transfer Learning on CIFAR10/100. }
\begin{center}
\fontsize{8pt}{12pt}\selectfont
\begin{tabular}{lcc}
 \toprule
 Models &CIFAR10&CIFAR100  \\
 \midrule
   \oursabbrv-4-384&95.54 &79.96   \\
  \oursabbrv-8-384&96.64 &82.09   \\
 \oursabbrv-8-512&97.03  &83.83   \\
\bottomrule
\end{tabular}
\end{center}
\label{tab:transfer_learning}
\vspace{-4mm}
\end{table}

\end{document}